\documentclass{article}
\usepackage{ijcai26}

\usepackage{pdfpages}
\usepackage{times}
\usepackage{soul}
\usepackage{url}
\usepackage[hidelinks]{hyperref}
\hypersetup{
    colorlinks=true,   % 开启文字颜色，去掉难看的方框
    citecolor=green,   % 参考文献引用的颜色 (如 \cite)
    filecolor=magenta, % 本地文件链接的颜色
    urlcolor=magenta      % 外部网址的颜色 (如 \href 和 \url)
}
\usepackage[utf8]{inputenc}
\usepackage[small]{caption}
\usepackage{graphicx}
\usepackage{amsmath}
\usepackage{amsthm}
\usepackage{booktabs}
\usepackage{algorithm}
\usepackage{algorithmic}
\usepackage[switch]{lineno}
\usepackage{named}
\usepackage{amsfonts}
\usepackage{setspace}
\usepackage[dvipsnames]{xcolor}

\title{{VibeFlow}: Versatile Video Chroma-Lux Editing through Self-Supervised Learning}

\author{
    Yifan Li$^1$
    \and
    Pei Cheng$^2$\and
    Bin Fu$^2$\and
    Shuai Yang$^{1}$\footnote{Corresponding Author.}\and
    Jiaying Liu$^1$\\
    \affiliations
    $^1$Wangxuan Institute of Computer Technology, Peking University
    $^2$Tencent\\
    \emails
    liyifan02@stu.pku.edu.cn,
    peicheng@tencent.com,
    fubin@gmail.com,\\
    \{williamyang, liujiaying\}@pku.edu.cn \\    
    \vspace{1.5mm}
    \href{https://lyf1212.github.io/VibeFlow-webpage/}{\texttt{https://lyf1212.github.io/VibeFlow-webpage/}} \\
}

\begin{document}

\maketitle

\begin{figure*}[t]
    \centering
    \includegraphics[width=1.0\linewidth]{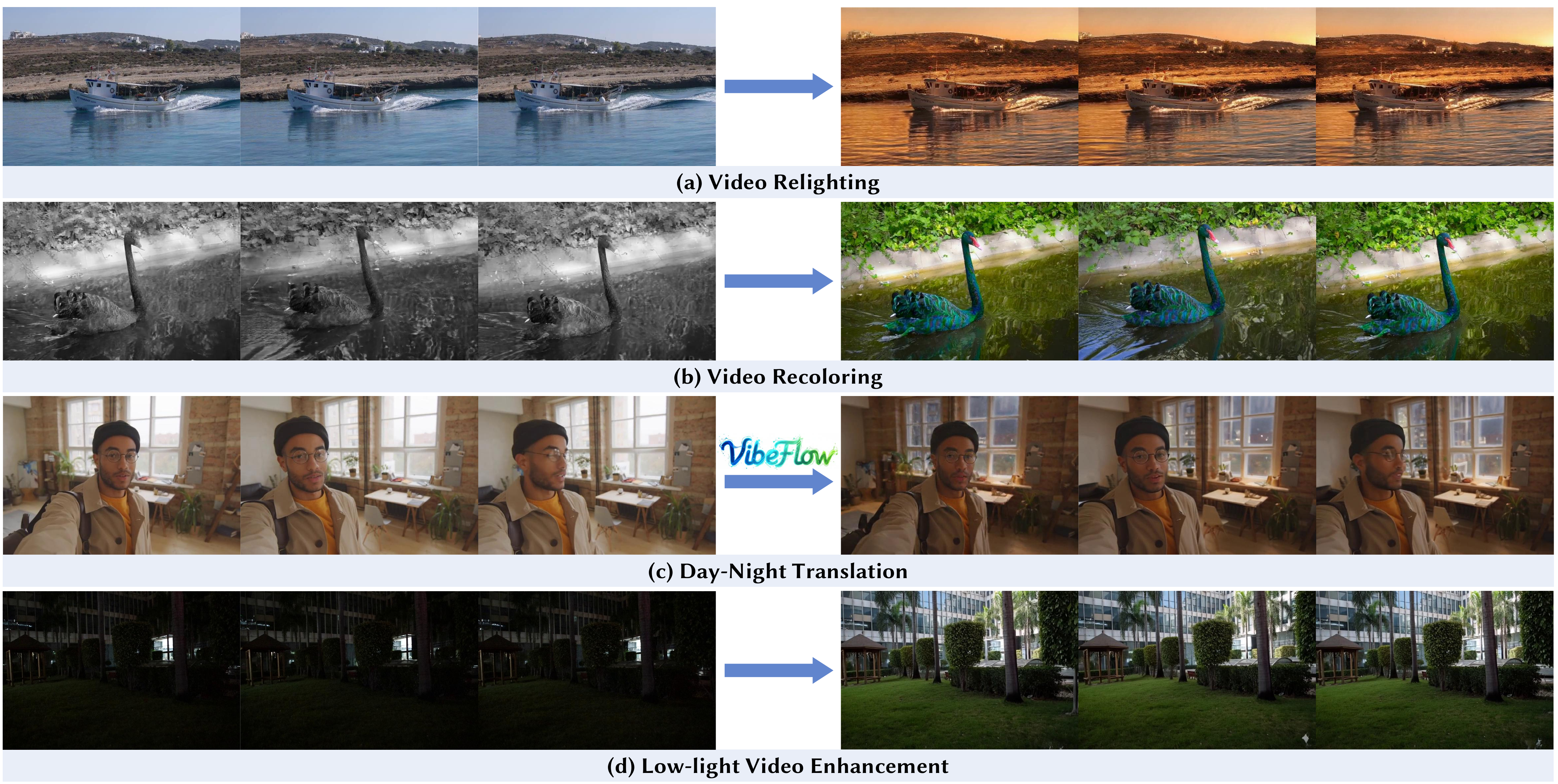}
    \caption{Our proposed \textbf{VibeFlow} achieves versatile video color and illumination editing while strictly preserve structural fidelity, enjoying rich applications, including video relighting, recoloring, day-night translation, low-light enhancement, and object-specific color editing.}
    \label{fig:teaser}
    \vspace{-4mm}
\end{figure*}

\begin{abstract}
    Video chroma-lux editing, which aims to modify illumination and color while preserving structural and temporal fidelity, remains a significant challenge. 
    %Existing methods typically rely on expensive supervised training with synthetic paired data, or suffer from compromised structural consistency and temporal coherence.
    Existing methods typically rely on expensive supervised training with synthetic paired data.
    %To address this, we propose
    This paper proposes \textbf{VibeFlow}, a novel self-supervised framework that unleashes the intrinsic physical understanding of pre-trained video generation models. 
    Instead of learning color and light transitions from scratch, we introduce a disentangled data perturbation pipeline that enforces the model to adaptively recombine structure from source videos and color-illumination cues from reference images, enabling robust disentanglement in a self-supervised manner.
    %without utilizing any paired data. 
    % Specifically, we selectively degrade structural details in reference images and color-illumination information in source videos, 
    Furthermore, to rectify discretization errors inherent in flow-based models, we introduce Residual Velocity Fields alongside a Structural Distortion Consistency Regularization, ensuring rigorous structural preservation and temporal coherence. 
    Our framework eliminates the need for costly training resources and generalizes in a zero-shot manner to diverse applications, including video relighting, recoloring, low-light enhancement, day-night translation, and object-specific color editing. Extensive experiments demonstrate that VibeFlow achieves impressive visual quality with significantly reduced computational overhead.
\end{abstract}

\section{Introduction}

The rapid evolution of diffusion models has significantly advanced the field of visual content editing. Within this landscape, effective control over \textbf{illumination and color} remains paramount, enjoying a wide range of high-value applications, including cinematic post-production~\cite{lc4,coco-lc}, photography retouching~\cite{pertouch}, and immersive environment design for embodied AI~\cite{tclight,relightmaster}.
% Precise control over illumination and color in a video fundamentally determines the realism, aesthetic appeal, and emotional tone, supporting a wide range of applications such as film production~\cite{lc4,coco-lc} and embodied AI simulation~\cite{tclight,relightmaster}.
Distinct from general video stylization~\cite{rerender} which requires proper modifications on texture details and geometry, \textbf{video chroma-lux editing necessitates strict temporal and structural fidelity} while synthesizing photorealistic colors and illumination.

Recently, large-scale unified editing models for both images~\cite{qwenimage,fluxkontext} and videos~\cite{vace,vino} have emerged, aiming to provide all-in-one editing solutions. While these models offer broad applicability, they struggle to maintain strict structural fidelity during editing~\cite{imgeditbench,vbench}, leading to inconsistent object identity and texture appearance unintentionally. 
% Furthermore, given the inherent high-dimensional complexity of video, these unified frameworks frequently face a generation-editing trade-off~\cite{colorturbo,vbench}, finding it difficult to simultaneously achieve high-quality editing and rigorous temporal consistency.

In contrast, specialized approaches tailored for color/light editing typically rely on task-specific synthetic datasets with supervised training. 
% As a pioneer in the field of image relighting,
For example, IC-Light~\cite{iclight} finetunes Stable Diffusion~\cite{sd} towards image relighting with over 10 million images.
% 
% collects over 10 million images with rendering pipeline or data augmentation, steering image generative backbones~\cite{sdxl,sd} towards scalable and physically plausible relighting.
% Some methods~\cite{synthlight} adopt intrinsic decomposition methods for explicitly disentangled image illumination editing.
CtrlColor~\cite{CtrlColor} and COCO-LC~\cite{coco-lc} both utilize ControlNet~\cite{controlnet} and paired gray/color data for image recoloring.
Although effective, these methods predominantly focus on image chroma-lux editing, paying less attention to video applications that present a more critical need.

To bridge this gap, existing methods can be categorized into \textbf{three} groups. The \textbf{first} one \textbf{extends} image diffusion models to the video domain by incorporating cross-frame mechanisms for improved coherency, such as temporal attention modules~\cite{tclight,relightvid}, feature inflation~\cite{animatediff}, and inter-frame object binding strategy\cite{lc4}, reducing temporal flickering and fluctuation.
However, they still struggle to generate natural and coherent videos as they learn temporal consistency from scratch.
The \textbf{second} category represents \textbf{training-free} approaches~\cite{anyv2v,anyportal,lvmin25flowportal}, which aim to alleviate the expensive training overhead, however still limited by the lack of video-specific optimization and handcrafted priors, producing unsatisfactory results.
% By leveraging temporal feature fusion and high-frequency information injection, these methods directly adapt image relighting approaches~\cite{iclight} to video sequences.
% However, limited by the lack of video-specific optimization and the reliance on handcrafted priors, these methods often produce temporally incoherent and visually implausible results.
% 
The \textbf{third} category involves \textbf{finetuning} video generation models ~\cite{relightmaster,uniLumos,lumen} on specific synthesized data. However, these methods rely heavily on auxiliary rendering data or physical priors (e.g., normal maps, albedo, depth) to supervise the training process, which is computationally expensive and labor-intensive, thereby restricting these methods to specific scenes. 
In summary, such specialized methods are typically confined to narrow applications and require dedicated training.

Building upon these observations, we identify that the fundamental challenge of video chroma-lux editing lies in simultaneously satisfying three critical objectives: \textbf{temporal coherence, photorealistic color/light manipulation, and strict structural fidelity}. Existing approaches struggle to satisfy these demands without incurring prohibitive training/inference costs or relying on scarce paired data.

Fortunately, content creation has witnessed a major transformation thanks to the rapid emergence of large-scale video generation models.
In contrast, inspired by~\cite{videolearner}, our approach stems from a key intuition: \textit{\textbf{since modern video generation models can synthesize realistic and continuous videos with complex lighting and colors, they must inherently possess a profound understanding of these photometric properties.}} 
Driven by this insight, we introduce \textbf{VibeFlow}, a novel self-supervised framework with a disentangled data perturbation pipeline, \textbf{mitigating the need for external physical supervision or paired data}.

Specifically, we propose an innovative self-supervised proxy task for improved structural fidelity.
Based on the video generation backbones that conditioned on a source video and a reference image, we selectively perturb specific components within the conditional inputs and finetune the model to reconstruct the original video, reinforcing the video generation models to learn the implicit disentanglement and adaptive recomposition of structural and color/light components from the conditional source videos and reference images respectively. 
Without any paired data, our method achieves versatile and precise modulation of photometric attributes, specifically illumination and colors, while rigorously preserving the temporal consistency and fine-grained details of source videos.

With the disentangled framework established, an intriguing question naturally arises: \textit{If we utilize the first frame of the source video as the reference condition, can the model faithfully reconstruct the input video?}
While theoretically implying an identity mapping, we empirically observe structural degradation due to the discretization error of flow-based solvers~\cite{sd3}.
To mitigate this issue, we propose the \textbf{Residual Velocity-guided Structural Preservation}.
We define the Residual Velocity Fields to rectify the sampling trajectory, which is conceptually inspired by~\cite{lvmin25flowportal,flowportal}.
Serving as the structural details compensation, the proposed residual velocity is derived from the reconstruction deviation under the original color/light conditions, which is subsequently added to the predicted velocity field of the target conditions, effectively and efficiently rectifying the editing sampling trajectory.
Furthermore, to ensure these residual fields encapsulate pure structural information and minimally entangled with the original illumination or colors of source videos, we propose a \textbf{Structural Distortion Consistency Regularization}.
By enforcing consistent structural degradation across varying conditions during training, this regularization collectively improves the precise preservation of fine-grained details and motion dynamics.

In summary, our framework delivers a straightforward yet powerful approach to flexible video chroma-lux editing, while avoiding the reliance on costly paired data and minimizing training overhead. \textbf{Remarkably}, by fully unleashing the potential capabilities of the video generation backbone, we enable diverse applications with various image editing models~\cite{fluxkontext,quadprior++,qwenimage}, including video relighting, recoloring, day-night transfer, low-light enhancement, and object-specific color editing.
Extensive experiments show that our approach outperforms state-of-the-art methods in both visual quality and temporal consistency.
Our contributions can be summarized as follows:
\begin{itemize}
    \item We propose VibeFlow, a novel self-supervised framework for versatile video chroma-lux editing. Without the requirements of paired data or expensive training overhead, our approach achieves high-quality and high-fidelity video color-light editing.
    \item We define Residual Velocity Fields for improved structure preservation and temporal consistency. We first propose a sampling trajectory rectification mechanism, which is further empowered by our proposed structural distortion consistency regularization, collectively ensuring precise preservation of the source video's fine-grained details and motion dynamics.
    \item Building upon data perturbation pipeline and self-supervised disentangled training, our method demonstrates exceptional zero-shot generalization across a wide range of applications, including video relighting, recoloring, day-night translation, low-light enhancement, and object-specific color editing, furthering showcasing the versatility in real-world scenarios.

\end{itemize}

\begin{figure*}[ht]
    \centering
    \includegraphics[width=0.86\linewidth]{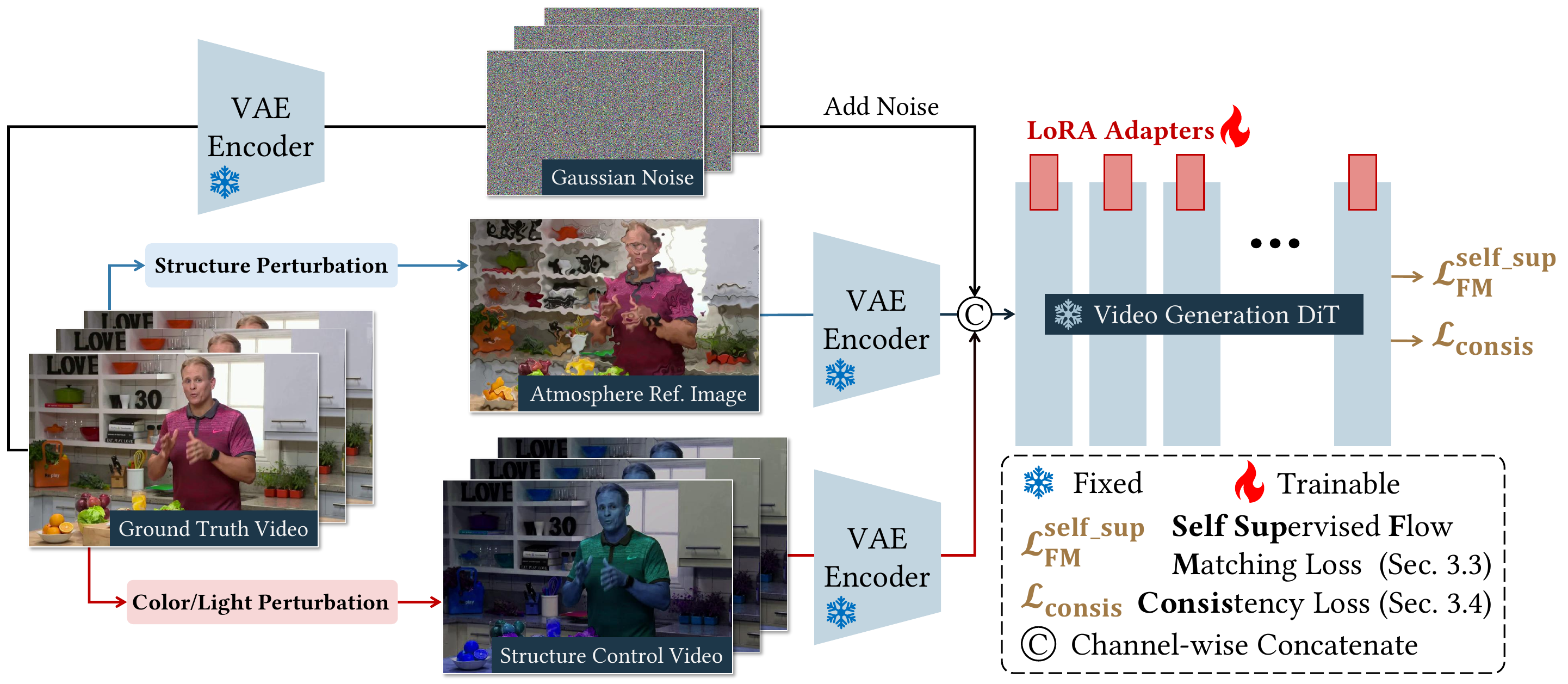}
    \caption{\textbf{Overall framework of our VibeFlow.} We propose a self-supervised training scheme with Structure/Chroma-lux data perturbation pipeline to finetune a video-to-video generation model using LoRA. Without any paired or rendered data, our approach achieves superior video chroma-lux editing with high-fidelity structural preservation.}
    \label{fig:pipeline}
    \vspace{-4mm}
\end{figure*}

\section{Related Works}
\subsection{Video Diffusion Models}
Inspired by the success of diffusion models in 2D image generation, recent advances in video diffusion models~\cite{svd,blattmann2023align,yang2024cogvideox,wan2025wan} have enabled the temporally coherent video generation conditioned on text, images, and richer auxiliary guidances.
Recently, CogVideoX~\cite{yang2024cogvideox} and Wan-series~\cite{wan2025wan} embrace the paradigm shift towards Diffusion Transformer (DiT) architectures, unifying video sequences and multi-modal conditions (e.g., text and images) into a scalable, tokenized representation, achieving more realistic and coherent video generation.
While these models excel at synthesizing plausible and temporally consistent videos, they inherently lack precise controllability, struggling to preserve fine-grained structural details during video-to-video translation. To address this limitation, our framework rigorously maintains structural fidelity and enables versatile video chroma-lux editing.

\subsection{Video Editing}
\noindent \textbf{Video Recoloring.} 
Early approaches utilize temporal convolutions~\cite{deepremaster} for temporal consistency, bidirectional feature fusion~\cite{bistnet}, or generative adversarial networks (GANs) for improved realism and consistency~\cite{vcgan}.
DeepRemaster~\cite{deepremaster} utilize temporal convolutions for reference-based video colorization, however, it is limited by poor generative capability, resulting in desaturated or unrealistic colors. VCGAN~\cite{vcgan} incorporates generative adversarial networks (GANs) for improved realism and consistency, however limited by unstable training, leading to unexpected artifacts.
More recently, L-C4~\cite{lc4} extends image diffusion models to video by training additional temporal consistency modules and inference-time optimization.
However, these methods are limited to broader applications.
Besides, without leveraging video diffusion models, they fail to capture a holistic understanding of the video sequences, leading to poor inter-frame consistency.

\noindent \textbf{Video Relighting.}
Video relighting remains a formidable challenge, requiring both physical realism and temporal consistency.
As a pioneer in the field of image relighting, IC-Light~\cite{iclight} collects over 10 million images with rendering pipeline or data augmentation, steering image generative backbones~\cite{sdxl,sd} towards scalable and physically plausible relighting.
Based on this, some of existing methods adapt IC-Light to video domain with temporal designs~\cite{tclight,anyportal,relightvid,lightavideo}.
To harness this generative power for video chroma-lux editing, recent methods~\cite{lumen,uniLumos,relightmaster} finetune existing video generation models~\cite{yang2024cogvideox,wan2025wan} using massive paired datasets generated via virtual engines. 
For instance, UniLumos~\cite{uniLumos} incorporates explicit physical supervision, utilizing depth and normal maps as auxiliary losses to bridge the domain gap.
While yielding promising results, these methods require laborious data synthesis and prohibitive training overhead (\textit{e.g.,} 8 NVIDIA H20 GPUs) to explicitly learn the relighting mapping from scratch.
Furthermore, all these approaches require costly training on large amounts of data.
In contrast, our approach mitigates all these requirements. 
By avoiding reliance on external rendering priors, we effectively unlock the superior capability of large-scale video generation backbones to understand and edit color and illumination in in-the-wild videos.

\noindent \textbf{General Video Editing.}
Towards consistent video editing, recent works like TokenFlow~\cite{tokenflow} AnyV2V~\cite{anyv2v} and Rerender A Video~\cite{rerender} have explored temporal propagation during per-frame editing. While effective for artistic stylization, these methods often prioritize structural changes over the photometric manipulation required for video chroma-lux editing. Our work bridges this gap by focusing on high-fidelity structure preservation without compromising the visual plausibility.

\section{VibeFlow}
\subsection{Preliminary}

\noindent \textbf{Flow Matching.} Typical DDPM~\cite{ddpm} facilitates an iterative noise-addition and denoising process to characterize real world image distribution.
Recently, SD3~\cite{sd3} represents a paradigm shift from DDPM to Rectified Flow~\cite{rectifiedflow}, enabling straight sampling trajectories between data and noise distributions, reducing error accumulation and enabling high-quality generation with fewer sampling steps:
\begin{equation}
    x_t=(1-t)x_0+t\epsilon, \quad t\in[0,1], \quad\epsilon\sim \mathcal{N}(0,\mathbf{I}).
\end{equation}
Then the denoising network $\mathbf{v}_\theta$ is trained with objective $\mathcal{L}_\text{FM}$ to predict the correct velocity pointing from noise $\epsilon$ to clean data $x_0$:
\begin{equation}
    \mathcal{L}_\text{FM}=\mathbb{E}_{x_0, t,\epsilon\sim \mathcal{N}(0,\mathbf{I})}\left[\lambda(t)\cdot||\mathbf{v}_\theta(x_t, t, c)-(\epsilon-x_0)||_2\right],
\end{equation}
where $\lambda(t)$ denotes the reweighting scheme based on the timestep $t$, $c$ indicates auxiliary conditions.
After training, any ODE Solver (such as Euler Solver) can be applied to sample a trajectory $\{x_t\}_{t=0}^{t=1}$, resulting in the sampled images or videos $x_0$:
\begin{equation}
\label{eq:ode}
    x_{t-\Delta t}\leftarrow\text{Solver}(x_t, \mathbf{v}_\theta(x_t,t,c)), 
\end{equation}
where $\Delta t$ denotes discretized timestep interval.

\subsection{Overview}
As illustrated in Fig.~\ref{fig:pipeline}, our framework utilizes LoRA~\cite{lora} to finetune a video-to-video generation model $F$ with a self-supervised training pipeline, demonstrating superior structural preservation capability on video chroma-lux editing.
Given a source video $\{I_\text{src}^i\}_{i=1}^{N}$ with $N$ frames and a reference image $I_\text{ref}$ generated by any image editing model, our \textbf{VibeFlow} achieves versatile video color-light editing without any paired data or costly training\footnote{We omit prompt conditions here and set them to ``high quality lighting and colors" during training and inference.}:
\begin{equation}
    \{I_\text{tgt}^i\}_{i=1}^{N} = F(\{I_\text{src}^i\}_{i=1}^{N}, I_\text{ref}),
\end{equation}
where $\{I_\text{tgt}^i\}_{i=1}^{N}$ indicates the edited videos.
Our key designs are two-fold: (1) \textbf{disentangled self-supervised training scheme}, taming video generation backbone to adaptively merge low-frequency colors/lights and high-frequency structural details, and (2) \textbf{residual velocity-guided structure preservation}, as introduced as follows.

\subsection{Disentangled Self-supervised Training Scheme}
\label{sec:ssl}
Since paired data of the same scene under different lighting conditions is limited,
we introduce a simple yet effective self-supervised training scheme that can be applied to arbitrary video datasets, as depicted in Fig.~\ref{fig:consis_loss}.
We design a disentangled data perturbation pipeline, and enforce the video generation model to reconstruct the original video: (1) low-frequency color/illumination perturbation $\mathcal{T}_\text{low}$ to be applied on source videos $\{I_\text{src}^i\}_{i=1}^{N}$, and (2) high-frequency structural information perturbation $\mathcal{T}_\text{high}$ to be applied on the reference image $I_\text{ref}$.
Formally, the self-supervised training loss can be expressed as:
\begin{align}
    \mathcal{L}&_\text{FM}^{\text{self-sup}}=\mathbb{E}_{x_0, t,\epsilon\sim \mathcal{N}(0,\mathbf{I})}\nonumber\\
    &\left[||\mathbf{v}_{\hat{\theta}}\left(x_t, t, \mathcal{T}_\text{low}(\{I_\text{src}^i\}_{i=1}^{N}), \mathcal{T}_\text{high}(I_\text{src}^1)\right)-(\epsilon-x_0)||_2\right],
\end{align}
where $\hat{\theta}$ denotes the LoRA-finetuned parameters of the video generation model, $x_0=\{I_\text{src}^i\}_{i=1}^{N}$.
By forcing the model to reconstruct the original video from these corrupted inputs, the network has successfully learned to \textbf{selectively extract structural information exclusively from conditional video, and color-illumination cues solely from the reference image for adaptive recomposition.}
In practice, we adopt the combination of elastic transformation and gaussian blur as $\mathcal{T}_\text{high}$, and color jitter as $\mathcal{T}_\text{low}$. 
Figure~\ref{fig:pipeline} provides a visualized example.
More details are in our supplementary materials.
\subsection{Residual Velocity-guided Structure Preservation}
\begin{figure}
    \centering
    \includegraphics[width=0.7\linewidth]{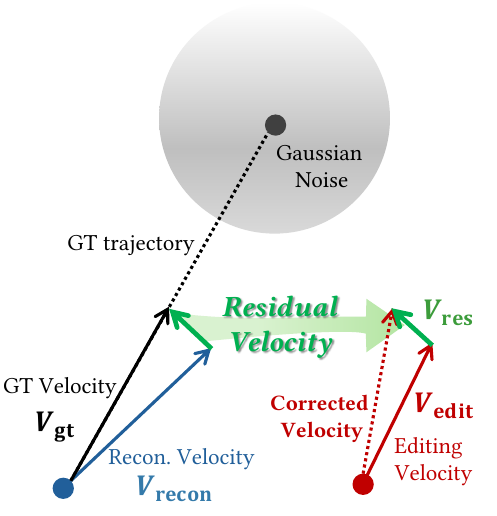}
    \caption{\textbf{Residual Velocity-guided Trajectory Rectification.} We define the residual velocity field to serve as structural compensation, correcting the editing sampling process iteratively.}
    \label{fig:res_velocity}
    \vspace{-2mm}
\end{figure}

\noindent \textbf{Residual Velocity Fields.} 
Given the flow-based nature of flow matching models~\cite{sd3,wan2025wan}, a fundamental discrepancy exists: models are optimized over continuous time space, yet inference using discrete solvers with skipped steps. 
This inherently introduces discretization errors even in reconstruction, which are further amplified under diverse conditions during video editing. 
Drawing inspiration from~\cite{lvmin25flowportal,flowportal}, we define the residual velocity field $V_\text{res}^t$ to serve as a structural compensation. This field is derived from the reconstruction deviation under source conditions:
\begin{align}
\label{eq:res_v}
    V_\text{gt}&=\epsilon-x_0,\nonumber\\
    V_\text{recon}^t &= \textbf{v}_{\hat{\theta}}(x_t, t, \{I_\text{src}^i\}_{i=1}^{N}, I_\text{src}^{(1)}), \nonumber\\\
    V_\text{res}^t&:=V_\text{gt}-V_\text{recon}^t.
\end{align}

\noindent \textbf{Trajectory Rectification.}
During inference, the residual velocity field acts as a geometric anchor to rectify the editing trajectory conditioned on target color/light references. 
Let $V_\text{edit}^t = \mathbf{v}_{\hat{\theta}}(x_t, t, \{I_\text{src}^i\}_{i=1}^{N}, I_\text{ref})$ denote the target editing velocity conditioned on the reference image $I_\text{ref}$ for the targeted color/light cues. 
Based on Eq.~(\ref{eq:ode}), the rectified trajectory at timestep $t$ is reformulated as:
\begin{equation}
    x_{t-\Delta t} \leftarrow \text{Solver}(x_t, V_\text{edit}^t + \gamma \cdot V_\text{res}^t),
\end{equation}
where $\gamma$ is a scaling factor controlling the strength of structural compensation and is set to 1, $\Delta t$ denotes discretized timestep interval.
This operation effectively injects the fine-grained structural details from the source videos into the edited ones, compensating for the inherent discretization errors without altering the target color/lights.
In practice, we skip the calculation in some timesteps and use the intermediate cache of $V_\text{res}$ to further boost the inference.

\noindent \textbf{Structural Distortion Consistency Regularization.}
\begin{figure}
    \centering
    \includegraphics[width=0.99\linewidth]{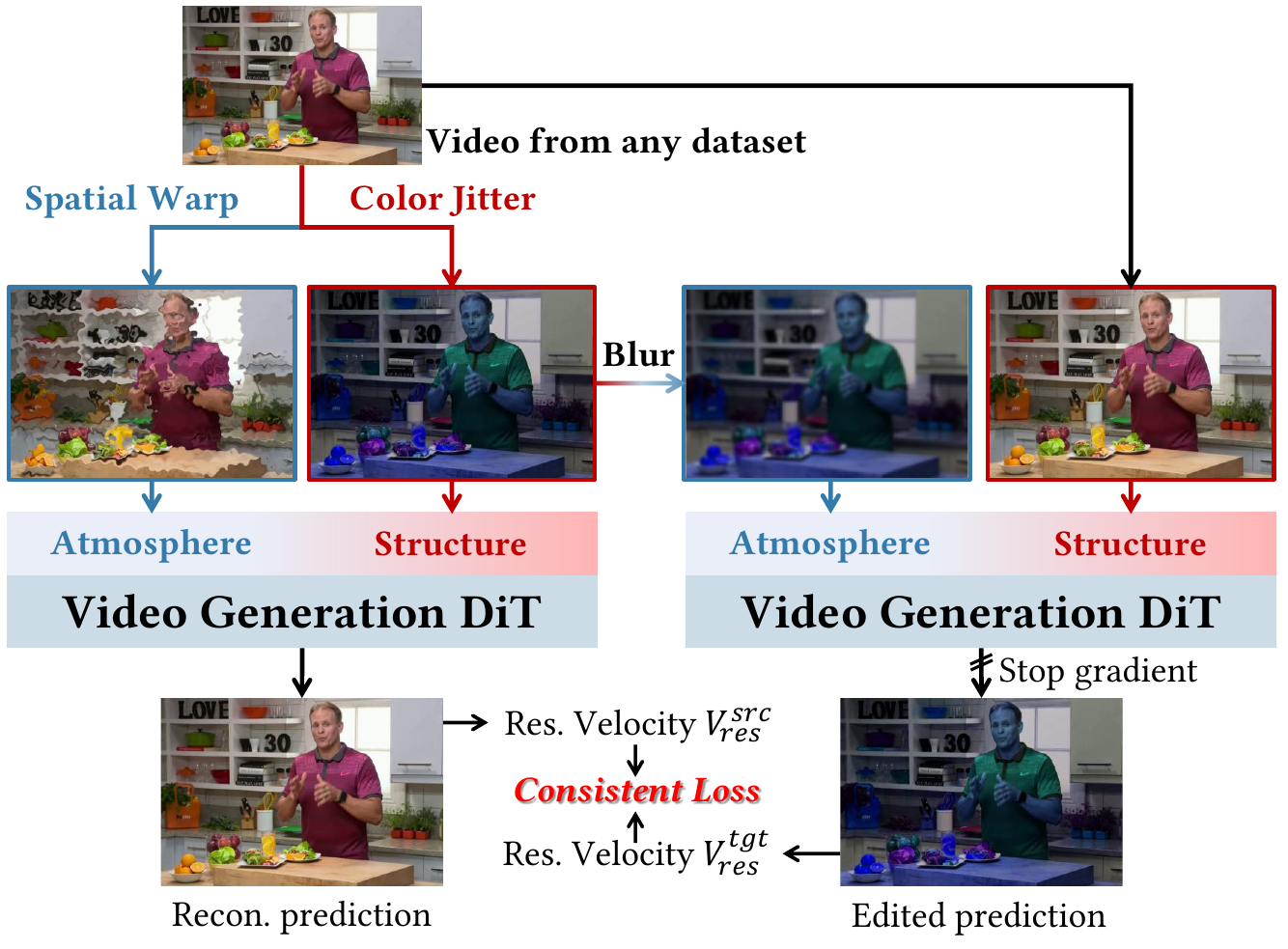}
    \caption{\textbf{Structural Distortion Consistency Regularization}, which performs a dual-branch training scheme.
    The left branch conducts the primary disentangled training under source conditions, while in the right branch, we swap the source and perturbed conditions, aiming at structural degradation-invariant capability.
    A stop-gradient operation is applied to the right branch, preventing the network from learning to synthesize the perturbed samples.}
    \label{fig:consis_loss}
    \vspace{-3mm}
\end{figure}
While the residual velocity $V_\text{res}$ is successfully designed to compensate for structural discretization errors, naively transferring the source-derived residual to the target condition is problematic.
The discretization errors are sometimes affected by different color/light conditions, leading to incompatible residual velocities across source and target conditions and incurring \textit{\textbf{structural misalignment and color leakage}}.

To address this limitation and further empower residual velocity-guided trajectory rectification, we propose a Structural Distortion Consistency Regularization strategy during training, as illustrated in Fig.~\ref{fig:consis_loss}.
Our objective is to enforce the model's structural degradation to be invariant to color/light variations, thereby ensuring that the residual velocity remains robust and transferable across different conditions.
In detail, we construct a dual-branch training scheme with swapped color/light conditions, and minimize the discrepancy of residual velocities.
Formally, given a source video $x_\text{src}$, we first get its augmented version $x_\text{aug}$ using the color/light perturbation $\mathcal{T}_\text{low}$ mentioned in Sec.~\ref{sec:ssl}.
Paralleled with the source-conditioned reconstruction branch that takes source video as atmosphere cues, we replace color/light conditions with gaussian-blurred perturbed data $\mathcal{G}(x_\text{aug}^{(1)})$, and utilize $x_\text{src}$ as structural conditions.
We then enforce a consistency loss between the residual velocities computed from these two branches:
\begin{align}
    V_\text{src}^\text{res} &= (\epsilon-x_\text{src}) - \textbf{v}_{\hat{\theta}}(x_t, t ,x_\text{aug}, I_\text{src}^{(1)})\nonumber,\\
    V_\text{aug}^\text{res} &= (\epsilon-x_\text{aug}) - \textbf{v}_{\hat{\theta}}(x_t, t ,x_\text{src}, \mathcal{G}(x_\text{aug}^{(1)}))\nonumber,\\
    \mathcal{L}_\text{consis} &= \mathbb{E}_{t, \epsilon \sim \mathcal{N}(0,\mathbf{I})} \left[ \| V_\text{src}^\text{res} - \text{sg}(V_\text{aug}^\text{res}) \|_2^2 \right],
\end{align}
where $\text{sg}(\cdot)$ denotes the stop-gradient operation, $\mathcal{G}$ is a gaussian blur kernel.
Based on this regularization, we enforce the network $\textbf{v}_{\hat{\theta}}$ to filter out atmospheric dependencies from the residual velocity, ensuring that $V_\text{res}$ solely focuses on compensating for structural fidelity and motion dynamics.

\subsection{Training Objectives}
The final training loss of our VibeFlow can be formulated as:
\begin{equation}
    \mathcal{L}=\mathcal{L}_\text{FM}^{\text{self-sup}} + \alpha \cdot \mathcal{L}_\text{consis},
\end{equation}
where $\alpha$ denotes the weighting hyper-parameter.

\begin{figure*}[ht]
    \centering
    \includegraphics[width=0.99\linewidth]{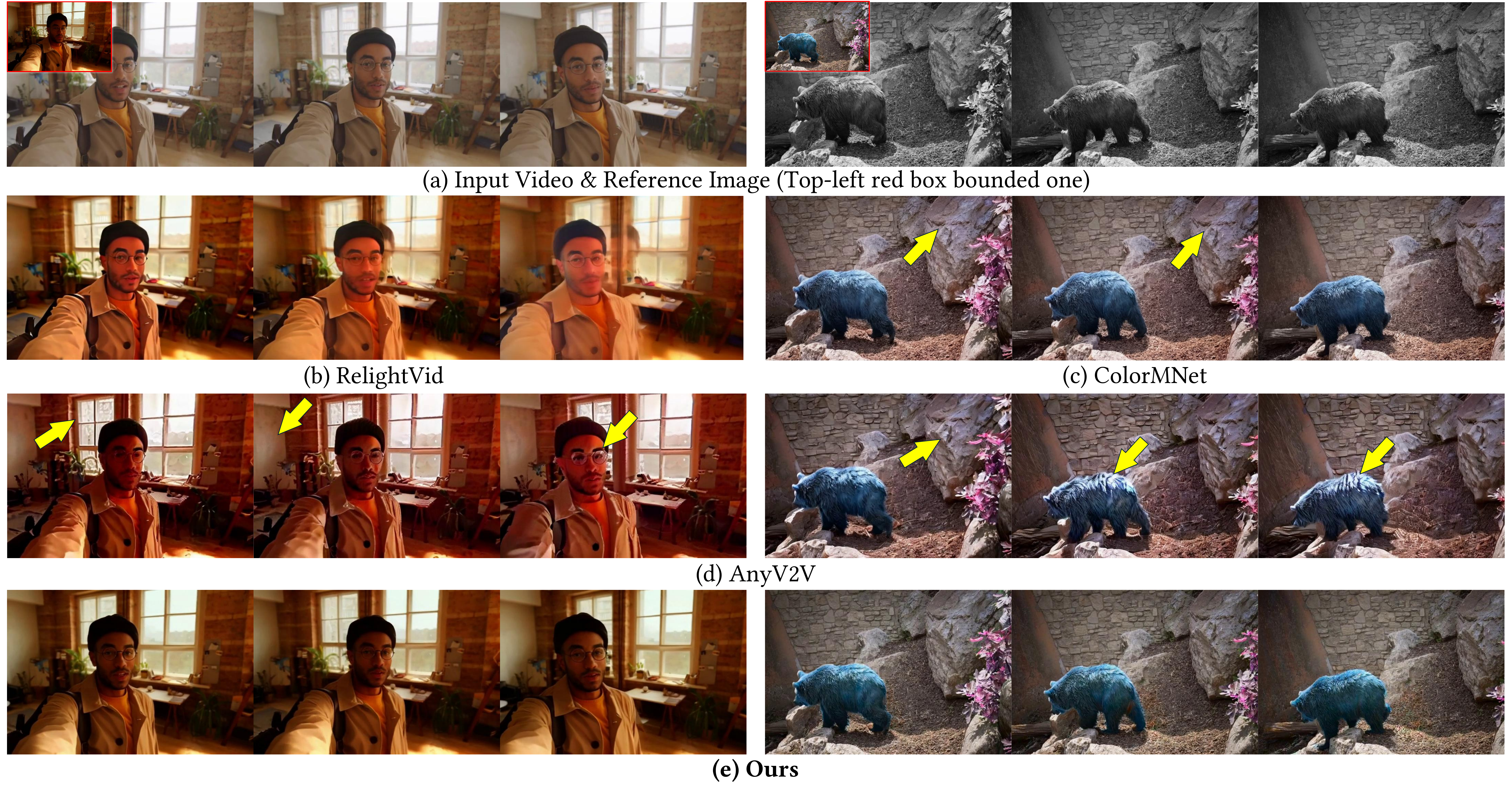}
    \caption{\textbf{Visualized Comparison.} Our VibeFlow outperforms both specialized methods and general methods in terms of visual quality, temporal consistency and structural fidelity.}
    \label{fig:vis_main}
    \vspace{-3mm}
\end{figure*}

\section{Experiments}
\subsection{Implementation Details}
We implement our method based on Wan2.1-Fun-1.3B-Control~\cite{wan2025wan,videoxfun}.
\textbf{The training is highly efficient}, with batch size = 8 and 8 A100-40G GPU in 8 hours. We set LoRA rank to be 128 and learning rate is 5e-5.
We adopt Qwen-Image-Edit~\cite{qwenimage} and Nano-Banana~\cite{nanobanana} to generate reference image based on the first frame of input video.
The overall pipeline takes 3-5 minutes to edit a video on an A100 GPU.

\noindent \textbf{Datasets.} We adopt 60 videos from DAVIS-val~\cite{davis2017val} as the evaluation set and 67520 samples from OpenVid-1M~\cite{openvid1m} as the training set.
% We first demonstrate our versatility on multiple video color-illumination editing tasks, including video relighting, recoloring, low-light enhancement, day-night transfer and object color editing. 

\noindent \textbf{Metrics.} We mainly adopt VBench~\cite{vbench} to calculate quantitative scores, including: (1) Subject Consistency \textbf{(SC)}: DINO~\cite{dinov1} MSE discrepancy between consecutive frames. (2) Structural Fidelity \textbf{(SF)}: SSIM~\cite{ssim} between Canny edge maps~\cite{controlnet} extracted from source and edited frames. (3) Temporal Coherency \textbf{(TC)}: Inter-frame CLIP~\cite{clip} similarity. (4) Motion Consistency \textbf{(MC)}: SSIM between two optical flow maps~\cite{raft}.

\subsection{Video Relighting and Recoloring}
\noindent \textbf{Comparison Methods.} We compare our VibeFlow with three specialized video relighting/recoloring methods: RelightVid~\cite{relightvid}, TC-Light~\cite{tclight}, ColorMNet~\cite{colormnet}, and two general video editing approaches: Wan2.1-1.3B~\cite{wan2025wan} and AnyV2V~\cite{anyv2v}.

\noindent \textbf{Qualitative Results.}
Fig.~\ref{fig:vis_main} shows visualized comparison on video relighting/recoloring.
RelightVid fails to fully preserve structural details, while AnyV2V produces unnatural artifacts and inconsistent textures as pointed by yellow arrows.
ColorMNet produces unsatisfactory color bias on the background rock.
In contrast, our method achieves superior visual quality and maintains temporal coherency and structural fidelity.

\noindent \textbf{Quantitative Results.}
We present metric scores in Tab.~\ref{tab:quantitative}. Our VibeFlow performs better than all comparison baselines in terms of all criteria, demonstrating versatile video chroma-lux editing while preserving strict structural fidelity.

\vspace{-2mm}
\begin{figure*}[t]
    \centering
    \includegraphics[width=0.99\linewidth]{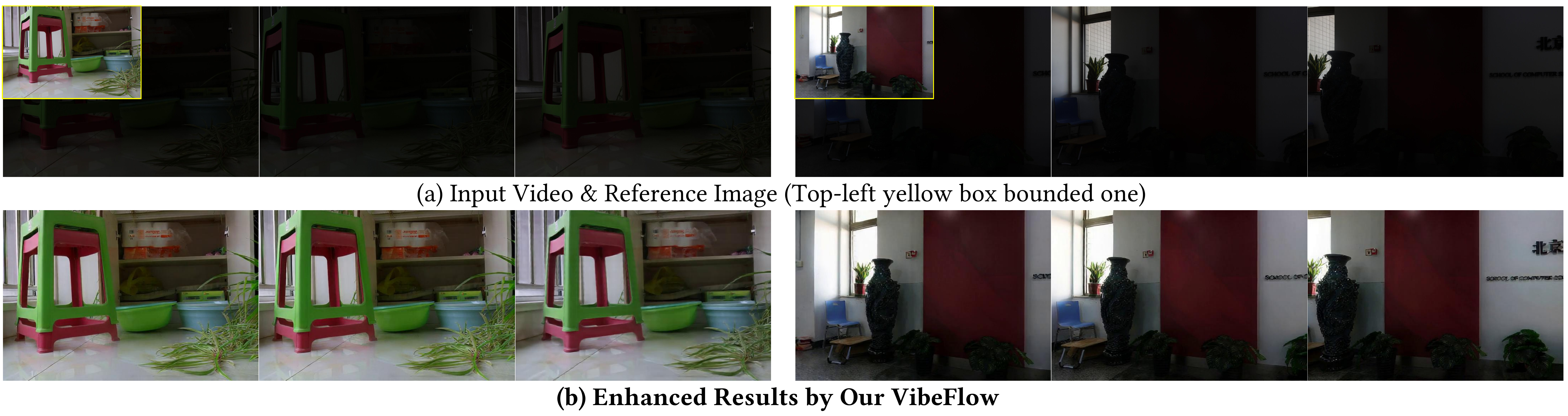}
    \caption{\textbf{Rich Applications.} Our VibeFlow can generalize to \textbf{video low-light enhancement} in a zero-shot manner with any specific training/finetuning, while demonstrating superior visual quality.}
    \label{fig:lowlight_vis}
    \vspace{-4mm}
\end{figure*}

\begin{table}[h]
    \centering
    % \scriptsize
    \footnotesize
    \caption{\textbf{Quantitative Comparison}, including four state-of-the-art specialized methods for relighting/recoloring, and two general methods. SC: Subject Consistency; SF: Structural Fidelity; TC: Temporal Coherency; MC: Motion Consistency.}
    \begin{tabular}{c|cccc}
    \toprule
        Method & SC & SF & TC & MC \\
        \midrule
        \multicolumn{5}{c}{\textit{Specialized Methods}} \\
         % TC-Light~\cite{tclight} & & & 0.3007 & 0.8311 \\
         TC-Light & 0.1226 & 0.9370 & 0.3007 & 0.8311 \\
         % RelightVid~\cite{relightvid} & & & 0.2638 & 0.8291 \\
         RelightVid & 0.2322 & 0.8668 & 0.2638 & 0.8291 \\
         % ColorMNet~\cite{colormnet} & & & 0.3021 & 0.8923 \\
         ColorMNet & 0.1310 & 0.9201 & 0.3021 & 0.8923 \\
         \midrule
         \multicolumn{5}{c}{\textit{General Methods}} \\
         Wan-2.1-1.3B & 0.1703 & 0.8329 & 0.2896 & 0.8586 \\
         AnyV2V~ & 0.1628 & 0.8710 & 0.2803 & 0.8647 \\
         \textbf{Ours} & \textbf{0.0817} & \textbf{0.9402} & \textbf{0.3061} & \textbf{0.9028} \\
    \bottomrule
    \end{tabular}
    \label{tab:quantitative}
    \vspace{-3mm}
\end{table}

\subsection{Rich Applications}
\begin{figure}[t]
    \centering
    \includegraphics[width=1.0\linewidth]{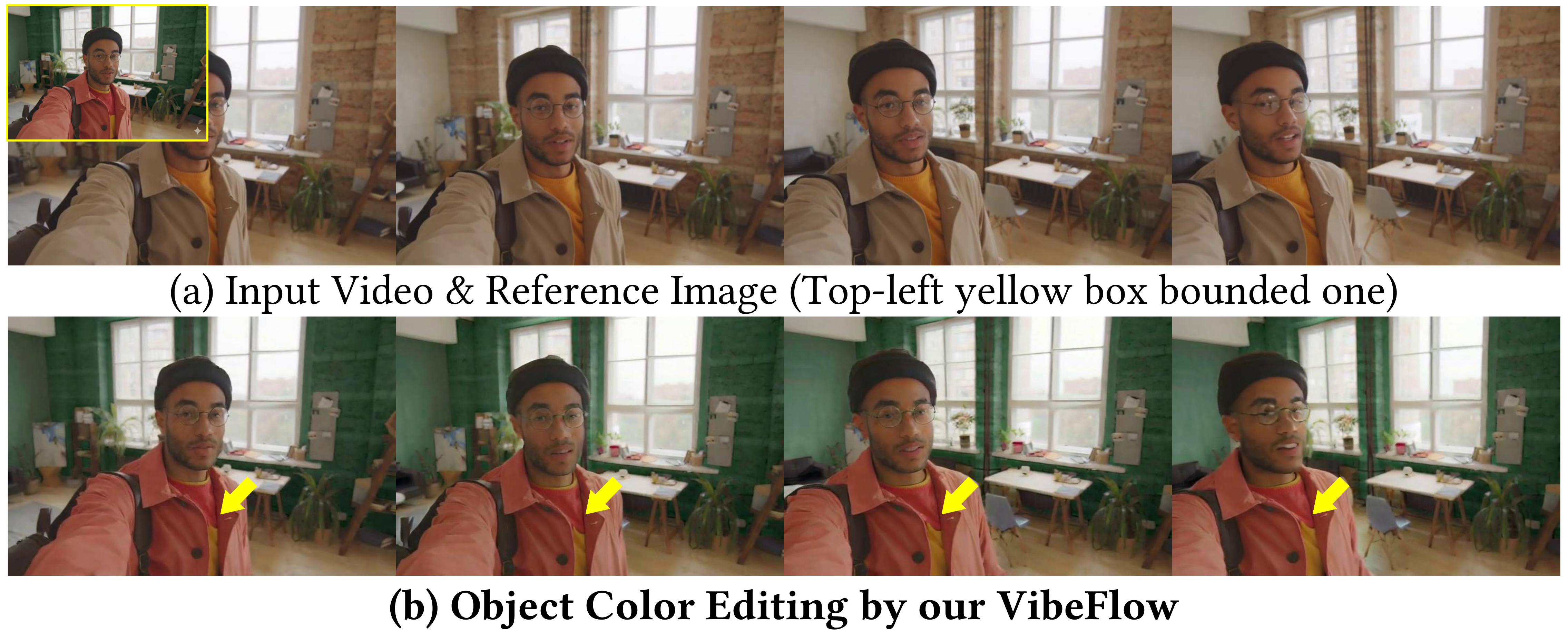}
    \caption{\textbf{Rich Applications.} Our VibeFlow can zero-shot generalize to \textbf{object color editing}, handling complex colors replacement. \textbf{Notably}, the wrinkles pointed by arrows are faithfully preserved, and colors strictly adhere to the reference image.}
    \label{fig:color_edit}
    \vspace{-4mm}
\end{figure}

\noindent \textbf{Object Color Editing.}
As shown in Fig.~\ref{fig:color_edit}, our VibeFlow achieves versatile object color editing even with \textbf{complex} reference image. Notably, as pointed out by the yellow arrows, our method preserve the creases on clothes and edit the color correctly according to the reference, demonstrating our superior capability on structural fidelity and detail preservation.

\noindent \textbf{Video Low-Light Enhancement.}
We present more low-light enhancement results in Fig.~\ref{fig:lowlight_vis}.
With the disentangled self-supervised training, the model learns to extract meaningful structural information to reconstruct normal-light video guided by the reference image.
The enhanced videos are temporally coherent and visually plausible with rich detail reconstructed correctly.

\begin{figure}[h]
    \centering
    \includegraphics[width=0.99\linewidth]{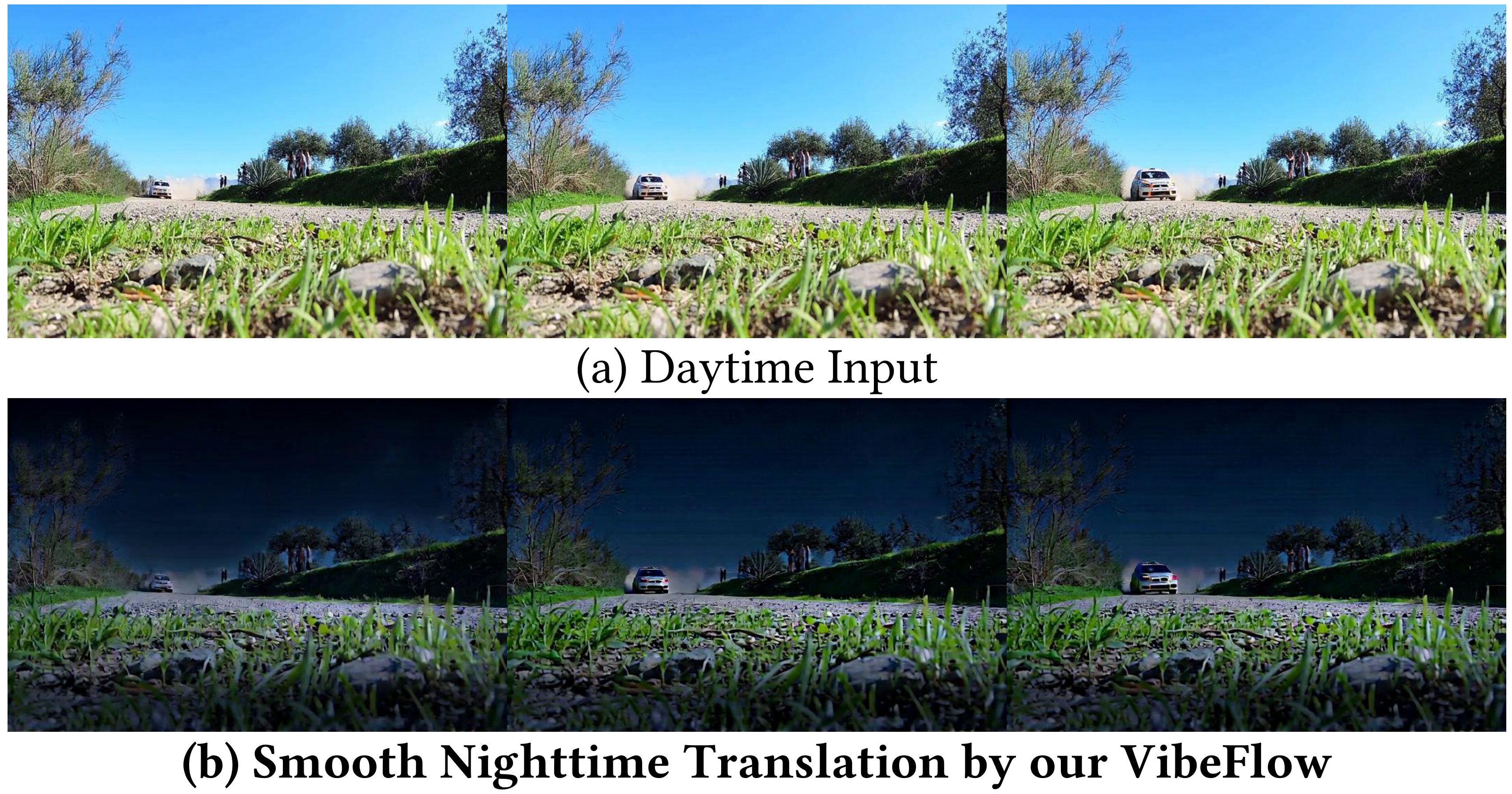}
    \caption{\textbf{Rich Applications.} Beyond basic color-light editing, our VibeFlow also achieves \textbf{day-night translation}, manipulating the colors of sky and illumination in a versatile manner.}
    \label{fig:nighttrans}
    \vspace{-2mm}
\end{figure}

\noindent \textbf{Video Day-Night Translation.}
Changing daytime scenes to nighttime requires not only editing on colors (sky from blue to black), but also darken the overall illumination.
With the powerful capability learned from disentangle training, the model can manipulate color and illumination intentively, achieving plausible day-night translation, as shown in Fig.~\ref{fig:nighttrans}.

\begin{figure}[h]
    \centering
    \includegraphics[width=0.99\linewidth]{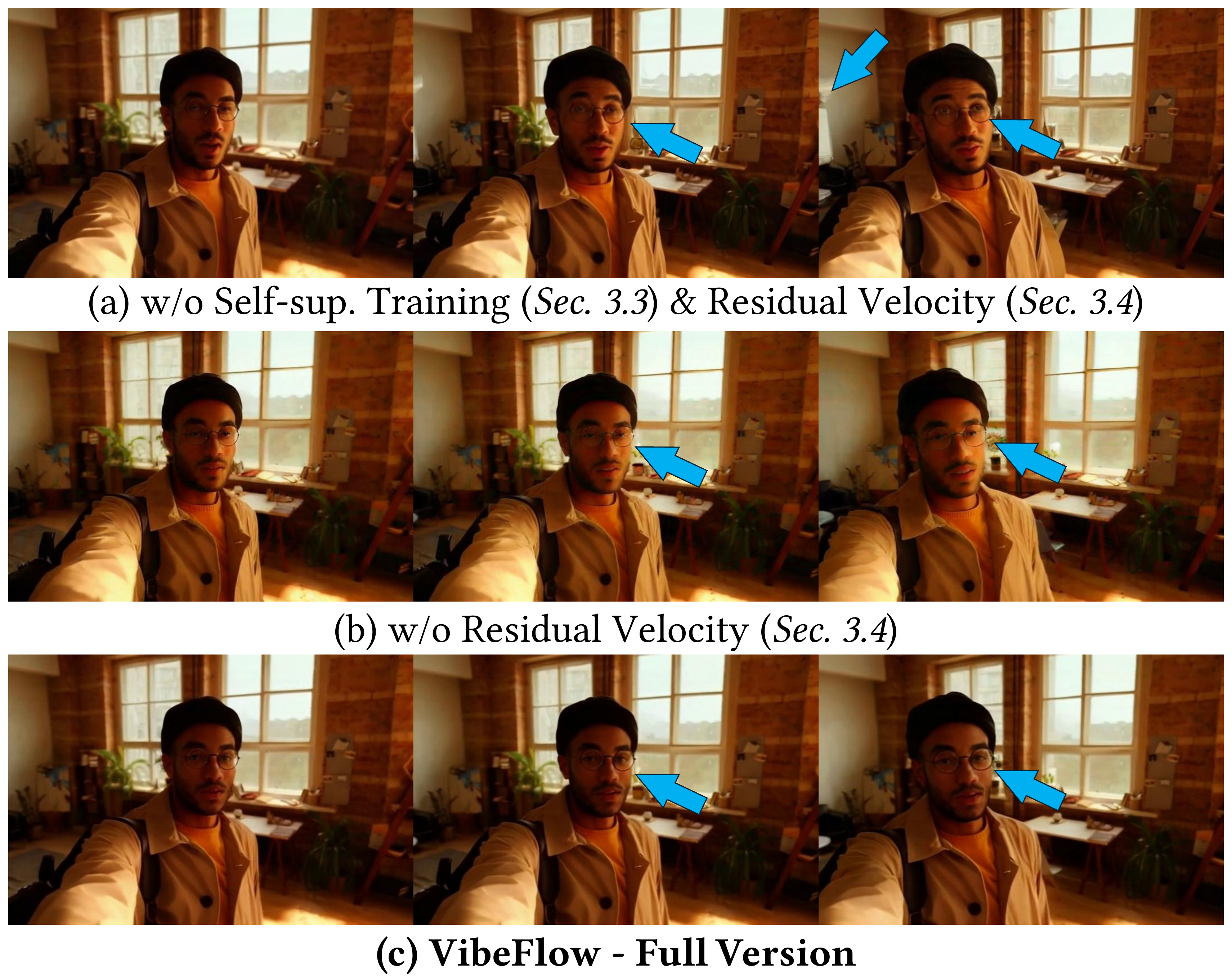}
    \caption{\textbf{Ablation Study.} \textbf{(Compare a\&c)} The lack of self-supervised disentangled training and residual velocity leading to severe chaotic eye movements and mismatched background sunlight. \textbf{(Compare b\&c)} Employing residual velocity-guided trajectory rectification and consistency training further alleviates the inherent discretization errors and improves fine-grained facial fidelity.}
    \label{fig:ablation}
    \vspace{-3mm}
\end{figure}

\vspace{-2mm}
\subsection{Ablation Study}

We perform ablation studies on our two key designs in Fig.~\ref{fig:ablation}.
% : (1) disentangled self-supervised training scheme, and (2) residual velocity-guided structural preservation.
Without our core designs, the model exhibits unnatural eye movements and discrepancies in the background sunlight.
While the self-supervised training effectively enforces global structural stability, it still limited by the inherent discretization error of original generative backbone, failing to preserve fine-grained facial details (e.g., eyes), as highlighted by the blue arrows.
With the proposed residual velocity-guided trajectory rectification and consistency training, the fine-grained facial details are preserved effectively.
% The ablation study verifies the effectiveness of our method.
\section{Conclusion}
% \section{Conclusion and Discussion}
In this paper, we propose VibeFlow, a self-supervised method for video color and light editing.
Unlike previous approaches, our method requires no paired data or costly training overhead. 
We fully unleash the powerful internal knowledge of pre-trained video generation models. 
Intuitively, we design a data perturbation strategy that forces the model to disentangle structure from atmosphere. 
To fix structural errors during generation, we define the residual velocity field and corresponding trajectory rectification, which is further empowered by our proposed consistency regularization scheme. 
Experiments demonstrate that VibeFlow achieves high-quality editing capability and can generalize to rich video chroma-lux editing tasks without further tuning.

\clearpage
\newpage
\begin{spacing}{0.99}
%% The file named.bst is a bibliography style file for BibTeX 0.99c
\bibliographystyle{named}
\bibliography{ijcai26}
\end{spacing}
\includepdf[pages={1-3}]{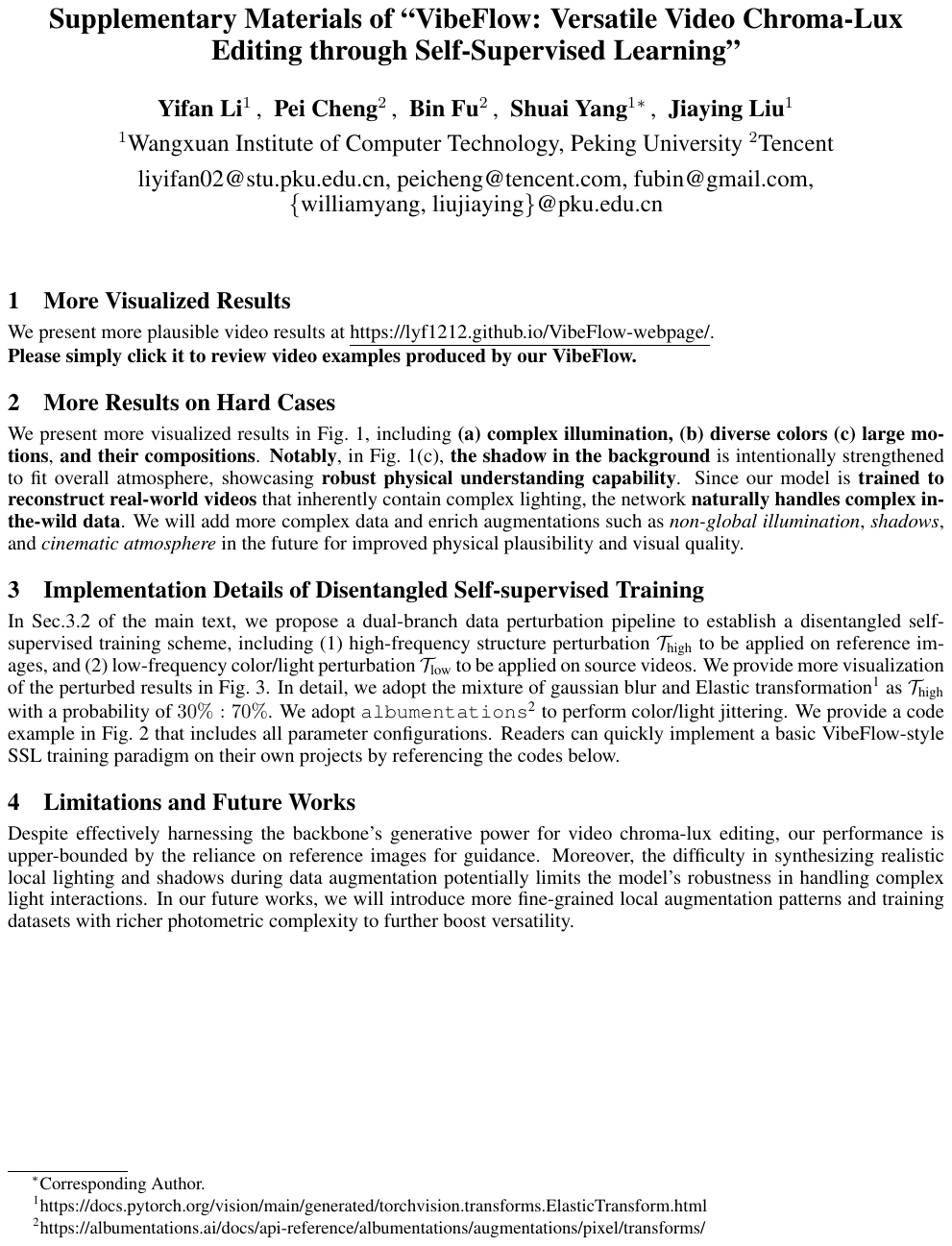}
\end{document}

\maketitle
% \onecolumn

\section{More Visualized Results}

We present more plausible video results with \underline{\texttt{video demo.html}} in our supplementary folder.

\noindent \textbf{Please simply click it to review video examples produced by our VibeFlow.}

\section{More Results on Hard Cases}
\begin{figure}[t]
    \centering
    \includegraphics[width=0.99\linewidth]{figs/hard cases.pdf}
    \caption{\textbf{Generalization clarification on hard cases.}}
    \label{fig:hard_cases}
\end{figure}

We present more visualized results in Fig.~\ref{fig:hard_cases}, including \textbf{(a) complex illumination, (b) diverse colors (c) large motions}, \textbf{and their compositions}.
\textbf{Notably}, in Fig.~\ref{fig:hard_cases}(c), \textbf{the shadow in the background} is intentionally strengthened to fit overall atmosphere, showcasing \textbf{robust physical understanding capability}. 
Since our model is \textbf{trained to reconstruct real-world videos} that inherently contain complex lighting, the network \textbf{naturally handles complex in-the-wild data}.
We will add more complex data and enrich augmentations such as \textit{non-global illumination}, \textit{shadows}, and \textit{cinematic atmosphere} in the future for improved physical plausibility and visual quality.

\section{Implementation Details of Disentangled Self-supervised Training}
\begin{figure}[h]
    \centering
    \includegraphics[width=1\linewidth]{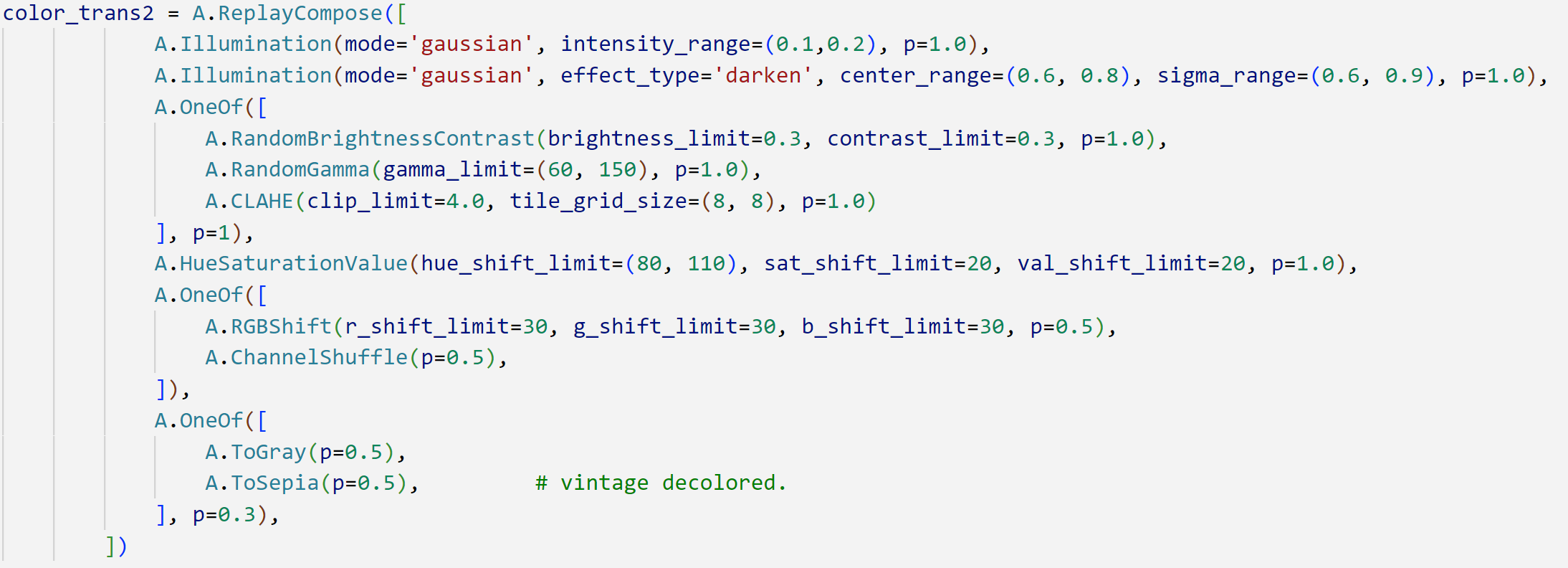}
    \caption{\textbf{Code example} that includes all parameter configurations used in color/light perturbation. Readers can quickly implement a basic VibeFlow-style SSL training paradigm on their own projects by referencing the codes above.}
    \label{fig:supp_color_jitter}
\end{figure}

\begin{figure}[t]
    \centering
    \includegraphics[width=0.99\linewidth]{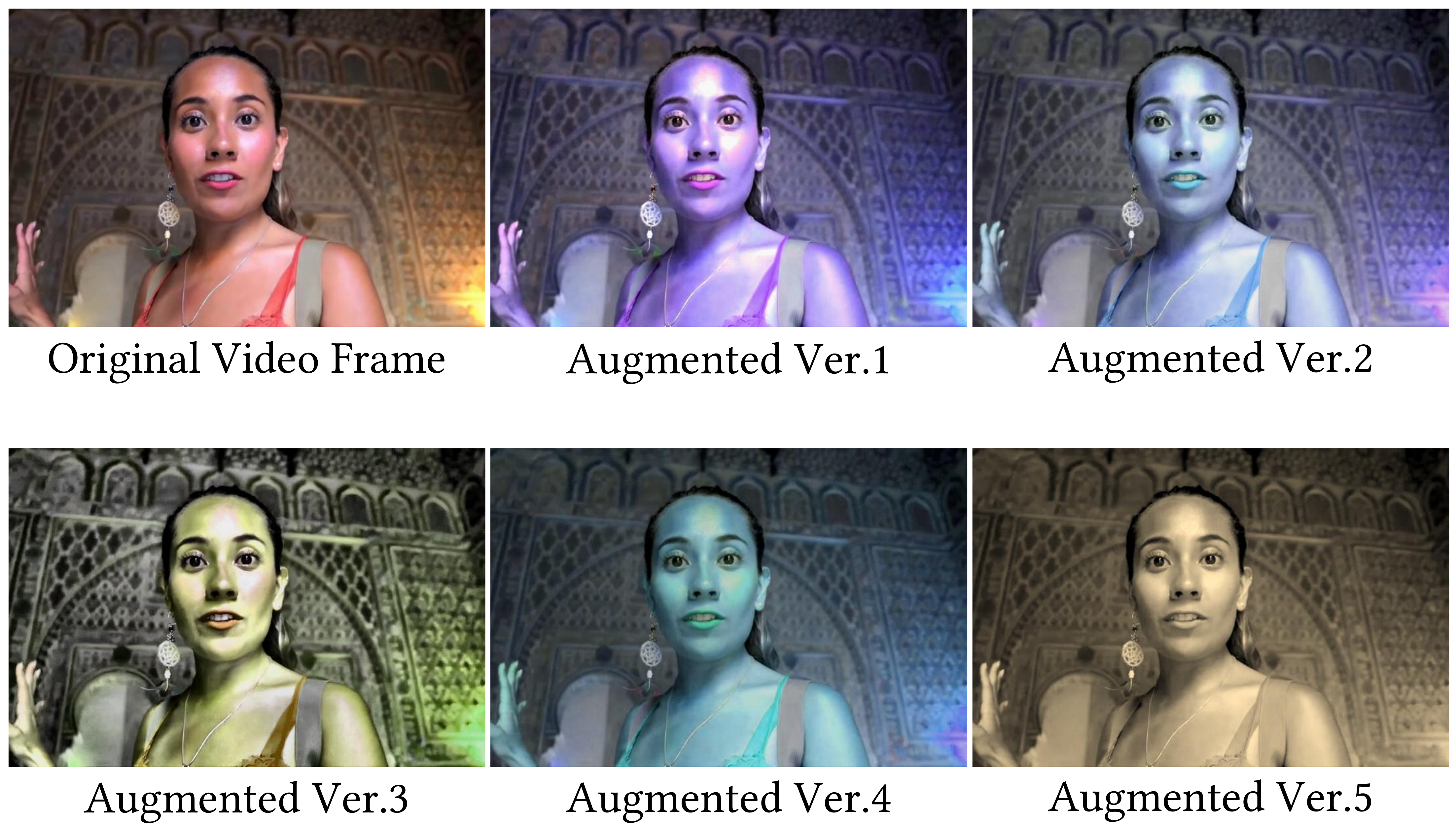}
    \caption{\textbf{Visualization of color/light perturbed results.}}
    \label{fig:supp_color_jitter_vis}
\end{figure}

In Sec.3.2 of the main text, we propose a dual-branch data perturbation pipeline to establish a disentangled self-supervised training scheme, including (1) high-frequency structure perturbation $\mathcal{T_\text{high}}$ to be applied on reference images, and (2) low-frequency color/light perturbation $\mathcal{T_\text{low}}$ to be applied  on source videos.
We provide more visualization of the perturbed results in Fig.~\ref{fig:supp_color_jitter_vis}.
In detail, we adopt the mixture of gaussian blur and Elastic transformation\footnote{https://docs.pytorch.org/vision/main/generated/torchvision.transforms.ElasticTransform.html} as $\mathcal{T_\text{high}}$ with a probability of $30\%:70\%$.
We adopt \texttt{albumentations}\footnote{https://albumentations.ai/docs/api-reference/albumentations/augmentations/pixel/transforms/} to perform color/light jittering.
We provide a code example in Fig.~\ref{fig:supp_color_jitter} that includes all parameter configurations.
Readers can quickly implement a basic VibeFlow-style SSL training paradigm on their own projects by referencing the codes below.

\section{Limitations and Future Works}
Despite effectively harnessing the backbone's generative power for video chroma-lux editing, our performance is upper-bounded by the reliance on reference images for guidance. Moreover, the difficulty in synthesizing realistic local lighting and shadows during data augmentation potentially limits the model's robustness in handling complex light interactions. 
In our future works, we will introduce more fine-grained local augmentation patterns and training datasets with richer photometric complexity to further boost versatility.
% \clearpage
% \newpage
% \begin{spacing}{0.943}
% %% The file named.bst is a bibliography style file for BibTeX 0.99c
% \bibliographystyle{named}
% \bibliography{ijcai26}
% \end{spacing}